\documentclass[sigplan]{acmart}
\usepackage{booktabs}
\usepackage{multirow}
\usepackage{tikz}
\usepackage{amsmath}
\usepackage[binary-units]{siunitx}
\usepackage[abbreviations]{foreign}
\usepackage{subcaption}
\usepackage{xspace}
\usepackage[inline]{enumitem}
\usepackage[algo2e,ruled,lined,boxed,commentsnumbered, noend, linesnumbered, procnumbered]{algorithm2e}
\usepackage{pifont}%

\usepackage{ifthen}
\newboolean{showcomments}
\setboolean{showcomments}{true}
\ifthenelse{\boolean{showcomments}}
{ \newcommand{\mynote}[3]{
		\fbox{\bfseries\sffamily\scriptsize#1}
		{\small$\blacktriangleright$\textsf{\emph{\color{#3}{#2}}}$\blacktriangleleft$}}
	\newcommand{\zzz}[1]{{\setlength{\fboxsep}{2pt}\fcolorbox{black}{yellow}{\textsf{\emph{#1}}}}\xspace}}
{ \newcommand{\mynote}[3]{}
	\newcommand{\zzz}[1]{}}

\usepackage{acronym}
\acrodef{DL}{decentralized learning}
\acrodef{ML}{machine learning}
\acrodef{D-PSGD}{decentralized parallel stochastic gradient descent}
\acrodef{FL}{federated learning}
\acrodef{SGD}{stochastic gradient descent}
\acrodef{IID}{independent and identically distributed}
\acrodef{non-IID}{non independent and identically distributed}
\acrodef{RMSE}{root mean square error}
\acrodef{RMW}{random model walk}
\acrodef{GL}{gossip learning}
\acrodef{EL}{epidemic learning}
\acrodef{DWT}{discrete wavelet transform}
\acrodef{FFT}{fast Fourier transform}
\acrodef{MI}{mutual information}
\acrodef{DP}{differential privacy}
\acrodef{VN}{virtual node}
\acrodef{RN}{real node}
\acrodef{LDP}{local differential privacy}
\acrodef{PNDP}{pairwise network differential privacy}
\acrodef{PNLDP}{pairwise network local differential privacy}
\acrodef{GI}{gradient inversion}
\acrodef{CML}{collaborative machine learning}
\acrodef{TPR}{true positive rate}
\acrodef{FPR}{false positive rate}

\acrodef{LA}{linkability attack}
\acrodef{GIA}{gradient inversion attack}
\acrodef{MIA}{membership inference attack}
\acrodef{AIA}{attribute inference attack}
\acrodef{ROC}{receiver operating characteristic}
\acrodef{AUC}{area under the ROC curve}
\acrodef{MoE}{mixture of experts}
\acrodef{LLM}{large language model}
\acrodef{HA}{Hardware Accelerator}
\acrodef{E2EIT}{end-to-end inference time}
\acrodef{TS}{tensor sharding}
\acrodef{GPU}{graphics processing unit}
\acrodef{TTFT}{time-to-first-token}
\acrodef{FFN}{Feed-Forward Network}
\acrodef{EP}{expert parallelism}
\acrodef{CF}{capacity factor}

\newcommand{\gpuset}{G\xspace}
\newcommand{\expertset}{E\xspace}

\newcommand{\batchsize}{b\xspace}
\newcommand{\seqlen}{s\xspace}
\newcommand{\hiddendimension}{h\xspace}

\newcommand{\hiddenin}{h_i} %
\newcommand{\hiddenout}{h_o} %

\newcommand{\routerskew}{\alpha_r\xspace} %
\newcommand{\numexpertskew}{k_r\xspace} %
\newcommand{\sys}{\textsc{MoEShard}\xspace}

\newcommand{\deepspeed}{\textsc{DeepSpeed}\xspace}
\newcommand{\pytorch}{\textsc{PyTorch}\xspace}
\newcommand{\bookcorpus}{\textit{BookCorpus}\xspace} %
\graphicspath{ {figures/} }
\usepackage{tikz}
\usepackage[eulergreek]{sansmath}
\usepackage{pgfplots}
\usepackage{pgfplotstable}
\usepackage{cleveref}
\usepackage{comment}
\crefname{assumption}{assumption}{assumptions}

\pgfplotsset{compat=newest}
\usepgfplotslibrary{external,units,colorbrewer,groupplots,fillbetween,statistics}
\tikzsetexternalprefix{figures/}
\tikzset{external/mode=list and make}
\usetikzlibrary{patterns,shapes.misc}

\makeatletter
\begingroup\endlinechar=-1\relax
\everyeof{\noexpand}%
\edef\x{\endgroup\def\noexpand\homepath{%
		\@@input|"kpsewhich --var-value=HOME" }}\x
\makeatother

\def\overleafhome{/tmp}
\newcommand{\inputplot}[2]{%
	\ifx\homepath\overleafhome%
	\IfBeginWith{#1}{plots}{\includegraphics{main-figure#2.pdf}}{#1}%
	\else%
	{\sffamily\scriptsize\input{#1}}
	\fi
}

\newcommand{\newgroupwidth}[2]%
{\expandafter\xdef\csname groupwidth#1\endcsname{#2}}

\newcounter{groupwidth}
\newsavebox{\groupwidthbox}
\makeatletter
{\edef\groupnumber{#1}%
	\stepcounter{groupwidth}%
	\@ifundefined{groupwidth\thegroupwidth}{\pgfmathsetlengthmacro{\mywidth}{\linewidth/\groupnumber}}%
	{\expandafter\let\expandafter\mywidth\csname groupwidth\thegroupwidth\endcsname}%
	\begin{lrbox}{\groupwidthbox}%
		\tikzset{/pgfplots/width={\mywidth}}%
		\ignorespaces}%
	{\end{lrbox}%
	\usebox\groupwidthbox
	\pgfmathsetlengthmacro{\mywidth}{\mywidth + (\linewidth - \wd\groupwidthbox)/\groupnumber}
	\immediate\write\@auxout{\string\newgroupwidth{\thegroupwidth}{\mywidth}}}
\makeatother
\usepackage{amsmath,amssymb,amsfonts,amsthm}
\usepackage{physics}
\usepackage{enumitem}
\usepackage{thm-restate}
\usepackage{bbm}
\theoremstyle{definition}

\theoremstyle{remark}

\allowdisplaybreaks


\begin{document}

\copyrightyear{2025}
\acmYear{2025}
\setcopyright{acmlicensed}\acmConference[EuroMLSys '25]{The 5th Workshop on Machine Learning and Systems }{March 30-April 3 2025}{Rotterdam, Netherlands}
\acmBooktitle{The 5th Workshop on Machine Learning and Systems (EuroMLSys '25), March 30-April 3 2025, Rotterdam, Netherlands}
\acmDOI{10.1145/3721146.3721938}
\acmISBN{979-8-4007-1538-9/2025/03}

\title[Accelerating MoE Model Inference with Expert Sharding]{Accelerating MoE Model Inference \\with Expert Sharding}

\author{Oana Balmau}
\affiliation{
  \institution{McGill University}
  \country{Canada}
  \city{Montreal}
}

\author{Anne-Marie Kermarrec}
\affiliation{
  \institution{EPFL}
  \city{Lausanne}
  \country{Switzerland}
}

\author{Rafael Pires}
\affiliation{
  \institution{EPFL}
  \city{Lausanne}
  \country{Switzerland}
}

\author{André Loureiro Espírito Santo}
\affiliation{
  \institution{EPFL}
  \city{Lausanne}
  \country{Switzerland}
}

\author{Martijn de Vos}
\affiliation{
  \institution{EPFL}
  \city{Lausanne}
  \country{Switzerland}
}

\author{Milos Vujasinovic}
\affiliation{
  \institution{EPFL}
  \city{Lausanne}
  \country{Switzerland}
}

\renewcommand{\shortauthors}{Balmau et al.}

\begin{abstract}

Mixture of experts (MoE) models achieve state-of-the-art results in language modeling but suffer from inefficient hardware utilization due to imbalanced token routing and communication overhead. 
While prior work has focused on optimizing MoE training and decoder architectures, inference for encoder-based MoE models in a multi-GPU with expert parallelism setting remains underexplored.
We introduce \textsc{MoEShard}, an inference system that achieves perfect load balancing through tensor sharding of MoE experts. 
Unlike existing approaches that rely on heuristic capacity factors or drop tokens, \textsc{MoEShard} evenly distributes computation across GPUs and ensures full token retention, maximizing utilization regardless of routing skewness.
We achieve this through a strategic row- and column-wise decomposition of expert matrices.
This reduces idle time and avoids bottlenecks caused by imbalanced expert assignments. 
Furthermore, \textsc{MoEShard} minimizes kernel launches by fusing decomposed expert computations, further improving throughput.
We evaluate \textsc{MoEShard} against \textsc{DeepSpeed} on encoder-based architectures, demonstrating speedups of up to 6.4$\times$ in time to first token (TTFT).
Our results show that when properly applied to experts, tensor sharding is a viable and effective strategy for efficient MoE inference.

\end{abstract}


\maketitle

\section{Introduction}
\label{sec:intro}
\acresetall

Scaling the size of \ac{ML} models has been a successful strategy to build generative \acfp{LLM}~\cite{fedus2022switchtransformers, palm-model-paper}.
These models are increasingly used in numerous domains such as healthcare and industry, and are becoming integral to modern society~\cite{bommasani2021opportunities}.
However, scaling these models introduces computational challenges and raises concerns about energy consumption and  sustainability~\cite{environment-llms}.

Conditional computation techniques can reduce the computational overhead during inference~\cite{moeoriginalpaper}.
\Acf{MoE} models implement conditional computation by replacing the feed-forward network in a transformer block by multiple smaller \textit{experts}.
Only a subset of experts (typically one or two) is activated per token input during inference.
A routing mechanism decides to which experts a particular token is forwarded.
This approach allows \ac{MoE} models to scale more efficiently than dense models.
However, these \ac{MoE} models have a significant memory footprint.
For example, the Switch-Base encoder-decoder model with 256 experts requires 54.63 GiB of memory, whereas the activated parameters of that model for one single token only requires 1.11 GiB.
Since a single \ac{GPU} often lacks the memory to store all experts, \ac{MoE} inference systems typically employ \emph{expert parallelism} where each \ac{GPU} holds a subset of experts~\cite{singh2023hybrid}.

While training \ac{MoE} models has received much attention in recent work~\cite{fast-moe-paper, faster-moe-paper, smart-moe-paper}, inference optimization remains under-explored.
A key challenge in \ac{MoE} inference with expert parallelism is the imbalance in workload distribution across \acp{GPU}~\cite{zhou2022expertchoicerouting, yang2021m6texploringsparseexpert,kim2024scalingbeyondtheGPU}.
Although routing mechanisms are trained to distribute tokens evenly among experts, in practice, some experts receive a disproportionate share of tokens, leading to uneven computational loads.
Moreover, this imbalance changes across different batches. %
This results in some \acp{GPU} idling while others remain fully utilized, increasing overall inference latency.
The end-to-end duration of inference is dictated by the \ac{GPU} with the most computational load (\eg, most tokens assigned), meaning that any load imbalance directly translates into inefficiencies in system throughput.

\begin{figure}[tpb]
	\centering
	\includegraphics{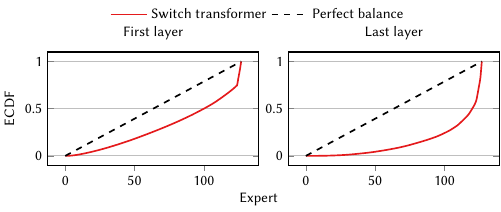}
	\caption{ECDF of token distribution per expert for the first and last layer, for a Switch transformer.
    }
	\label{fig:ecdf}
\end{figure}

We empirically show this imbalance in \Cref{fig:ecdf}.
We show an ECDF of token distribution per expert for the first and last layer of a Switch encoder-only model with 128 experts.
Particularly for the last layer, there are significant differences in the load on different experts.
For the last layer, 14 experts do not receive any token, whereas the most busy expert receives \num{3105} tokens.

Existing \ac{MoE} inference systems attempt to mitigate token imbalance through various strategies.
A common approach is to employ \acp{CF}, which limits the number of tokens assigned to each expert~\cite{zhou2022expertchoicerouting}.
However, this method often results in token dropping, which degrades model accuracy.
Other methods, such as expert replication, distribute copies of overburdened experts and tokens across multiple GPUs to balance the load~\cite{wang2023prophet,wu2024lazarus}.
While this alleviates some imbalance, it also requires profiling solutions and introduces additional overhead.
Thus, efficiently achieving a balanced workload across GPUs running \ac{MoE} models remains an open challenge.

This paper proposes \sys, an inference system that achieves perfect load balancing for \ac{MoE} models by applying \ac{TS} to experts.
In contrast to existing work, experts are not replicated, and no profiling is required.
Instead, our key insight is that the structure of the expert models and the associated computation is easily parallelizable across GPUs.
We, therefore, take advantage of the structure of the \ac{MoE} expert computation, which consists of a multiplication of two matrices.
This operation can be efficiently sharded (the first matrix column-wise, the second row-wise) so that each GPU holds a shard of each of the matrices for all experts.
Sharding like this achieves perfect load balancing as all the tokens can be processed in parallel for each batch.
Our work thus takes a novel way of looking at the load imbalance problem, in contrast to other approaches that alleviate load imbalance by replicating experts over multiple \acp{GPU} or redirecting tokens to different \acp{GPU}.

Our experiments compare \sys against \deepspeed, a state-of-the-art framework for distributed training and inference of large \ac{ML} models.
\sys achieves up to 6.4$\times$ speedups in terms of \ac{TTFT} and these speedups increase as the batch size grows.

This paper makes the following contributions.
\begin{itemize}
    \item We introduce \sys, a \ac{MoE} inference solution with perfect load balancing (\Cref{sec:design}).
    \sys achieves this by evenly distributing the expert computation across multiple \acp{GPU}.
    We minimize the computational overhead by grouping and fusing kernels.
    \item We implement \sys and conduct experiments, comparing the \ac{TTFT} latency of \sys against that of \deepspeed (\Cref{sec:evaluation}).
    Our experimental results show that \sys results in significantly lower \ac{TTFT} compared to \deepspeed and is a feasible approach to speed up \ac{MoE} model inference in token-imbalanced scenarios.
\end{itemize}

\section{Background on Multi-GPU MoE Inference}
\label{sec:prelims}

\textbf{Transformer models} have become a cornerstone of modern \ac{ML}~\cite{vaswani2017attention}. 
A transformer model comprises multiple transformer blocks, each leveraging self-attention mechanisms and \acfp{FFN} to process input tokens.
The self-attention mechanism enables the model to capture dependencies across the sequence by dynamically attending to different input elements. 
The resulting representations are then refined by a \ac{FFN}.

\textbf{Mixture-of-Experts (MoEs)} is a form of sparse computation where only a subset of specialized sub-networks, known as \textit{experts}, are activated during inference~\cite{moeoriginalpaper}.
In a \ac{MoE} model, certain transformer blocks can include an \ac{MoE} layer, which we refer to as a \emph{\ac{MoE} block}. 
Unlike a conventional transformer block with a single \ac{FFN}, a \ac{MoE} layer consists of multiple experts, typically between 8 to 256~\cite{shazeer2017outrageouslylargeneuralnetworks,fedus2022switchtransformers}.
Rather than propagating tokens to all experts, \ac{MoE} models dynamically \textit{route} each token to only a subset of experts.%

Since a single model generally cannot fit all experts within the memory of a single \ac{GPU}, parts of the model are processed by different \acp{GPU}.
To address this, \ac{MoE} model inference typically relies on \acf{EP}.
With \ac{EP}, the self-attention and router layers are replicated across GPUs, while the experts are distributed across GPUs~\cite{fedus2022switchtransformers}.
During the forward pass, each GPU processes a minibatch of input tokens independently, computing self-attention in parallel. 
The router on each GPU then assigns tokens from its minibatch to specific experts.
Since the assigned experts may reside on different GPUs, an all-to-all scatter communication step ensures that each GPU receives the tokens designated for the experts it hosts, introducing the first synchronization barrier in \ac{MoE} blocks. 
Once the tokens reach their respective GPUs, expert computations are performed locally using the assigned experts. 
Following this, an all-to-all gather communication step consolidates the computed results, returning them to the GPUs responsible for the original inputs. 
These processed tokens then serve as input for the next \ac{MoE} block.

\textbf{System assumptions.} Our work proposes a refinement of \ac{EP}. Instead of placing experts on each GPU, we \textit{shard} all experts and place pieces of each expert in each GPU, leveraging the parallelizable structure of expert computations.
In order to do this, our system makes the following assumptions:
\begin{enumerate*}[label=\emph{(\roman*)}]
\item All \acp{GPU} are considered to have equal computational capacity and memory;
\item the entire \ac{MoE} model fits in the collective memory of all GPUs;
\item we operate on a single server that hosts multiple GPU, all interconnected via high-speed, high-throughput links; and
\item we make the simplifying assumption that the number of expert shards is divisible by the number of GPUs, for the sake of clarity and space constraints. Handling scenarios with ``leftover'' shards is straightforward but remains outside the scope of this work.
\end{enumerate*}

\begin{algorithm2e}[t]
    \small
    \DontPrintSemicolon 
    \caption{\sys forward pass 
    }
    \label{alg:moeshard}
    \SetKwProg{Fn}{Procedure}{:}{end}
    \SetKwInOut{Input}{Require}
    \SetKwFunction{Forward}{forward}
    \SetKw{KwSet}{set}
    \Input{$\gpuset$: Set of GPUs, $\expertset$: Set of experts.}
    \Fn{\textup{\textsc{forward}}($ x $)} {
        \textbf{$ // $ Step 1: token routing} \;
        $ m_{expert} \leftarrow$ \textsc{router}($x$)\label{line:token_routing} \;
        \;

        \textbf{$ // $ Step 2: metadata exchange} \;
        $ I_{exp} \leftarrow $ \textsc{groupPerExpert}($x, m_{expert} $) \;
        $ m_{sizes} \leftarrow \textsc{countPerExpert}(I_{exp}) $ \;
        \textsc{sendMetadataToGPUs}($\gpuset$, $m_{sizes}$)\;
        \textbf{receive} $m'_{sizes}[g] $ from each GPU $ g \in \gpuset $\;
        \;
        
        \textbf{$ // $ Step 3: scatter tokens} \;
        \textsc{sendTokensToGPUs}($G$, $\textsc{concatenate}(I_{exp})$)\;
        \textbf{receive} $ W[g] $ from each GPU $ g \in \gpuset $ \;
        \;

        \textbf{$ // $ Step 4: expert computation} \;
        \For{$g \in \gpuset$} {
            \For{$e \in \expertset$} {
                $ s \leftarrow $ \textsc{loadShard}($g$, $ e $)\;
                $W[g][e] \leftarrow $\textsc{compute}($s$, $W[g][e]$)\;
            }
        }
        \;

        \textbf{$ // $ Step 5: gather tokens} \;
        \textbf{send} $ W[g] $ to each GPU $ g \in \gpuset $\;
        \textbf{receive} $ y[g] $ from each GPU $ g \in \gpuset $\;
        $ x \leftarrow $ \textsc{aggregateTokens}($ y $)\;
        \;
        \Return{x}\\
    }

\end{algorithm2e}

\begin{figure*}[t!]
	\centering
	\begin{subfigure}{.36\linewidth}
		\raisebox{1.2cm}{
			\includegraphics[width=\textwidth]{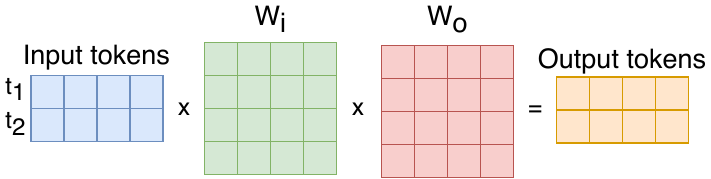}
		}
		\caption{Expert computation without parallelism.}
		\label{fig:expert_computation}
	\end{subfigure}%
	\hfill
	\begin{subfigure}{.58\linewidth}
		\includegraphics[width=\textwidth]{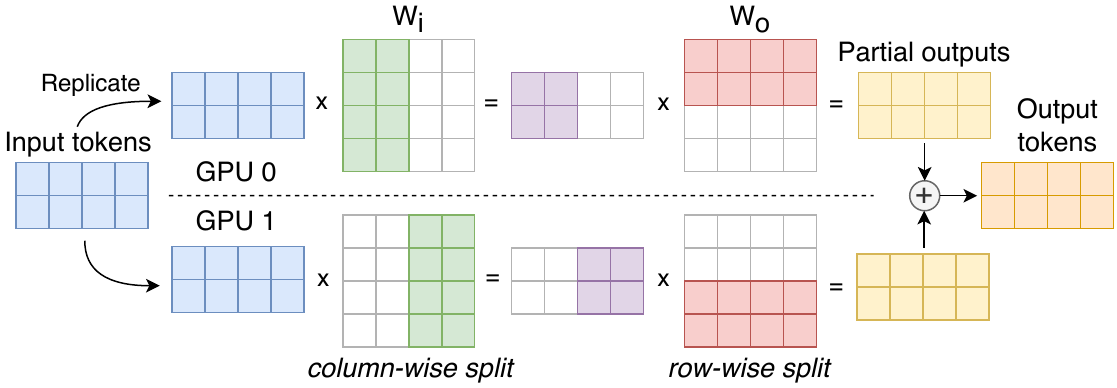}
		\caption{Expert computation with \sys and expert sharding.}
		\label{fig:expert_computation_slicing}
	\end{subfigure}
	\caption{Expert computations with and without \sys. An expert consists of matrices $ W_i $ (in green) and $W_o $ (in red).}
	\label{fig:split-matrix}
\end{figure*}

\section{Design of \sys}
\label{sec:design}

In a nutshell, with \sys, each GPU takes all tokens as input and hosts a shard of each expert to compute \emph{partial token outputs}.
These partial token outputs are combined later into a final output for each token.
All non-\ac{MoE} layers, as well as the components of \ac{MoE} layers excluding experts, are replicated across each GPU, following the work of Lepikhin et al.~\cite{g-shard-paper}. 

We first present the workflow of \sys in \Cref{sec:workflow}, then detail the expert sharding algorithm in \Cref{sec:slice-wi-wo}, and finally present an expert fusing optimization to minimize the computation overhead in \Cref{sec:optimize_expert_computation}.

\subsection{\sys workflow}
\label{sec:workflow}
\Cref{alg:moeshard} shows the pseudocode associated with the \textsc{forward} function where tokens are processed by the experts. 
We refer to the set of \acp{GPU} as $\gpuset$, and the set of \emph{experts} as $\expertset$.
Each \ac{GPU} executes the \textsc{forward} function on a tensor of input tokens $x$ with shape $[\batchsize, \seqlen, \hiddendimension]$, where $\batchsize$ represents the batch size, $\seqlen$ the sequence length, and $\hiddendimension$ the hidden dimension.
At this point, the self-attention layer has already processed these input tokens.
\sys operates in the following six steps:

\noindent\textbf{Step 1: token routing.}
The input tokens are first assigned to specific experts using a router mechanism by the \textsc{router} function.
This assignment creates a token-to-expert mapping, $ m_{expert} $, which is a tensor of integers representing the target expert for each token.

\noindent\textbf{Step 2: metadata exchange.}
This step ensures that each GPU knows the number of tokens assigned to each expert by every other GPU.
We first group each input token in $ x $ by its assigned expert as defined by the router, and store this grouping in $ I_{exp} $.
Then we create a list $ m_{sizes} $ of size $ |\expertset| $ where the value at each index $ i $ indicates the number of input tokens assigned to expert $ \expertset[i] $.
The list $ m_{sizes} $ of each GPU is sent to all other GPUs, and the per-GPU input counts are stored in $ m'_{sizes} $, completing the metadata exchange.

\noindent\textbf{Step 3: scatter tokens.}
\sys then replicates \emph{all} input tokens across all GPUs, \ie, each GPU sends its input tokens $ I_{exp} $ to all other GPUs, concatenating them into one tensor for communication efficiency.
The received tokens are stored by each GPU in a two-dimensional tensor $ W $.
Specifically, $ W[g][i] $ stores the tokens originating from GPU $ g $ designated for expert $e$. %
We note that each GPU can correctly map the incoming list of input tokens to entries in $ W $ using the input counts in $ m'_{sizes} $ received earlier.

As a consequence of our devised algorithm, all input tokens need to be replicated across all \acp{GPU}.
While this has implications for memory usage and communication volume, this overhead is manageable.
As an example, assume there are 4 \acp{GPU} with $\batchsize=\num{250}$, $\seqlen=\num{120}$, and $\hiddendimension=\num{768}$. Assuming a \SI{4}{\byte} occupation per tensor element, each \ac{GPU} must send approximately \SI{88}{\mebi\byte} to all other \acp{GPU} while receiving \SI{352}{\mebi\byte}.
Given that \textsc{NVLink} 3.0 supports up to \SI{600}{\gibi\byte\per\second} bidirectional bandwidth \cite{nvlink}, sending \SI{88}{\mebi\byte} per GPU would only take around \SI{0.15}{\milli\second}, which is negligible in the end-to-end inference time.

\noindent\textbf{Step 4: expert computation.}
\sys now processes the tokens in $ W $ by iterating over each GPU in $ \gpuset $ and expert in $ \expertset $.
We load the appropriate expert shard for each GPU $g$ and expert $e$ by calling the \textsc{loadShard} function.
The particular shard to load depends on the rank of the GPU that executes the \textsc{forward} function.
The relevant expert shard then processes the tokens, and the corresponding entries in $ W $ are replaced with the output of the expert computation.
Depending on the number of experts and GPUs, this results in many matrix multiplications, and we discuss how to optimize this step in \Cref{sec:optimize_expert_computation}.

\noindent\textbf{Step 5: gather tokens.}
The processed tokens in $ W[g] $ are then sent back to each GPU $ g $, and each GPU $ g $ receives the partial token outputs $ y[g] $.
These tokens are then point-wise aggregated, resulting in the final token outputs $ x $.
This aggregation is a consequence of our choice of expert sharding dynamics, which we elaborate on in the next subsection.

\subsection{Expert Sharding}
\label{sec:slice-wi-wo}
We now discuss how \sys shards experts across \acp{GPU}.
Experts are typically implemented as two matrices $ W_i \in \mathbb{R}^{(\hiddenin, \hiddenout)}$ and $ W_o \in \mathbb{R}^{(\hiddenout, \hiddenin)}$~\cite{fedus2022switchtransformers,g-shard-paper}.
We visualize a standard expert computation on the input tokens in \Cref{fig:expert_computation}, where the input tokens are processed using two matrix multiplications.
To balance the computational load across GPUs, \sys employs \emph{expert sharding} where the matrices $ W_i $ and $ W_o $ are split between different GPUs.
This allows the input tokens to be processed in parallel %
with the resulting partial computations aggregated to produce the final output.
The shards of one expert are contiguous rows and columns of $W_i$ and $W_o$, and each \ac{GPU} holds $\frac{a}{|G|}$ rows or columns of both matrices, where $a$ is either $\hiddenin$ or $\hiddenout$.
Each \ac{GPU} holds one shard of \emph{all} experts.

\Cref{fig:expert_computation_slicing} shows how an expert is split across two GPUs with \sys.
Matrix $ W_i $ is sharded column-wise, and $ W_o $ is sharded row-wise.
Thus, if matrix $ W_i $ has four columns ($\hiddenout = 4$), GPU 0 loads the first two columns, and GPU 1 loads the remaining two.
Similarly, if matrix $ W_o $ has four rows ($\hiddenout = 4$), GPU 0 loads the first two rows, and GPU 1 loads the remaining two.
Let $ W_i^g $ and $ W_o^g $ denote the shard of $ W_i$, respectively $ W_o $ held by GPU $ g $.
Each GPU $ g $ now computes $ x \cdot W_i^g \cdot W_o^g $, resulting in the partial output $ y_g $ with the same dimension as the input tokens $ x $.
Summing each $ y_g $ for each GPU $ g $ will yield equivalent output tokens as in \Cref{fig:expert_computation}.

We acknowledge that different sharding strategies are possible.
Generally, $ W_i $ and $ W_o $ can be sharded row-wise, column-wise, or in combinations of both.
We first analyze the sharding of $W_i$.
For simplicity, let $x$ have shape $(c, \hiddenin)$.
In a column-wise split, each GPU processes $c \cdot \hiddenin$ entries of $\mathbf{x}$. 
Conversely, in a row-wise split, each \ac{GPU} processes only $\frac{\hiddenin \cdot c}{|\gpuset|}$ entries of $\mathbf{x}$.
This distinction directly affects the volume of data transferred during the first data communication round.
With a column-wise split of $W_i$, each GPU must send all its input data to every other GPU, resulting in a total data transfer of $c \cdot \hiddenin \cdot (|\gpuset| -1)$ matrix entries per GPU. 
In contrast, a row-wise split requires each GPU to send only $\frac{c \cdot \hiddenin \cdot (|\gpuset| -1)}{|\gpuset|}$ entries in total, as each GPU transmits a unique segment of $\mathbf{x}$ to the others.

While a row-wise split of $W_i$ may initially seem advantageous, considering the interaction with $W_o$ reveals a different outcome. 
A column-wise split of $W_i$ allows both matrix multiplications to proceed without intermediate synchronization if $W_o$ is split row-wise.
All other sharding combinations would require synchronization between operations. 
For instance, if $W_i$ is split row-wise, the outputs of $\mathbf{x}W_i^P$ must be summed point-wise across all \acp{GPU}.
Similarly, if both $W_i$ and $W_o$ are split column-wise, the outputs of $\mathbf{x}W_i^P$ must be concatenated across \acp{GPU} before the next multiplication.

The optimal sharding strategy is to split $W_i$ column-wise and $W_o$ row-wise.
Assuming that $\hiddenout \equiv 0 \pmod{|\gpuset|}$, each \ac{GPU} will store $\hiddenin \cdot \frac{\hiddenout}{|\gpuset|}$ entries from  $W_i$  and $\frac{\hiddenout}{|\gpuset|} \cdot \hiddenin$ entries from $W_o$.

\subsection{Optimizing expert inference}
\label{sec:optimize_expert_computation}
\sys executes numerous small matrix multiplications as each GPU processes each expert shard independently.
This can result in substantial compute overhead due to the need for frequent kernel launches.
This overhead becomes particularly problematic as the number of experts and GPUs in the system increases.
To address this issue, we reduce the number of kernel launches using the following two optimizations.

Firstly, instead of separately processing the input tokens for each GPU and expert, we concatenate the tokens \emph{for the same expert from all GPUs} into a single tensor, thus reducing the maximum number of expert shard computations from $ |\expertset| \times |\gpuset| $ to $ |\expertset| $.
After the expert computations, we group back the tokens per GPU and assign them to the appropriate entries in $ W $.
This optimization makes the number of kernel launches for expert shard processing independent of the number of GPUs.

Secondly, \sys leverages variable-sized sparse matrix multiplication, enabling the processing of all expert shards in a single operation using a large sparse matrix multiplication algorithm, as detailed by Gale \etal~\cite{megablocks}.
This approach makes the number of kernel launches independent of the number of experts.
We empirically evaluate the effect of this optimization on performance in \Cref{sec:evaluation}.

\section{Evaluation}
\label{sec:evaluation}

We implement \sys in the Python 3 programming language using \pytorch\footnote{Source code is available at \url{https://github.com/sacs-epfl/moe-inference}.}.
We compare the performance of \sys against \deepspeed, a popular framework for \ac{MoE} inference.
Our experiments answer the following three questions:
\begin{enumerate}
    \item How does the per-layer inference latency of \sys compare to that of \deepspeed across different batches (\Cref{sec:exp_per_layer_latency})?
    \item How does the \ac{TTFT} of \sys evolve for \sys and \deepspeed when varying the number of experts and batch size (\Cref{sec:exp_speedup})?
    \item How does the \ac{TTFT} of \sys evolve with and without Sparse Matrix Multiplication when varying the number of experts and batch size (\Cref{sec:ablation_study})?
\end{enumerate}

\subsection{Experimental setup}

\textbf{Model and dataset.}
We evaluate \sys using Google Switch Transformers~\cite{fedus2022switchtransformers}, a family of language models that extend the T5 architecture~\cite{raffel2020exploring} by replacing its feed-forward layers with \ac{MoE} logic.
In particular, we use the Switch-Base version of the model.
Since autoregressive decoder generation is not particularly compute-intensive and relies more on fine-grained optimizations, we focus only on the encoder part of the model to understand the performance gains by our approach. %
All experiments are run on \bookcorpus, a large-scale dataset comprising up to \num{7185} unique books~\cite{zhu2015moviebook}.

\textbf{Router.}
To regulate skew in token-to-expert assignments, we replace the default router with a custom implementation, used in all experiments except \Cref{sec:exp_per_layer_latency}.
Since the Switch Transformer employs a \acf{CF}, our router ensures that all tokens are processed instead of being dropped. 
However, it is unsuitable for production due to its probabilistic nature, leading to nonsensical token-to-expert assignments. 
Nevertheless, it allows us to evaluate the performance of \sys under varying skew conditions.  

Our router has two parameters:
\begin{enumerate*}[label=\emph{(\roman*)}]
    \item \emph{router skew} ($\routerskew$), which controls token-to-expert imbalance, and  
    \item \emph{number of skewed experts} ($\numexpertskew$), the number of experts receiving a disproportionate share of tokens.  
\end{enumerate*}  
For each token, the router selects an expert from a multinomial distribution, where the selection probability $p_i$ of the expert indexed by $i \in [|\expertset|]$ is proportional to:
\[
p_i \propto \begin{cases}
    \frac{1}{|\expertset|} + \routerskew, & i \leq \numexpertskew \\
    \frac{1}{|\expertset|}, & \text{otherwise}
\end{cases}
\]

\textbf{Hardware.}
Our evaluation is executed using four NVIDIA A100 \acp{GPU}, each with \SI{80}{\giga\byte} GPU memory.
We use CUDA 12.6.
All \acp{GPU} are connected to the same computing node and share access to a single CPU, specifically an AMD EPYC 7543 32-core Processor operating at a maximum clock speed of 3.7 GHz.
The GPUs are interconnected via NVLink technology \cite{nvlink} and are linked to the CPU through PCIe bridges.

\textbf{Baseline.}
We compare \sys against \deepspeed, specifically \deepspeed-MoE, a popular inference engine for \ac{MoE} models~\cite{deepspeed-moe-paper}.
For a fair comparison, we enable expert parallelism in \deepspeed.
By default, \deepspeed employs a \ac{CF} in the router of \ac{MoE} layers. We fix this parameter to $\texttt{min(}|\expertset|\texttt{, 50)}$ to minimize token loss, as we found increasing the \ac{CF} further leads to memory issues.
Notably, \deepspeed also implements expert sharding; however, its purpose is to scale the system horizontally rather than to address load balancing.
In both \deepspeed and \sys, the non-expert parts of the model are replicated across all \acp{GPU}.

\subsection{Per-layer latency of \sys and \deepspeed}
\label{sec:exp_per_layer_latency}

We compare the per-layer latency of \sys and \deepspeed by measuring the average forward-pass latency across multiple encoder layers. Using a batch size of 250, a sequence length of 120, and 128 experts, we collect results over 100 iterations and average them per layer. This experiment employs the default, original router.  

Our measurements show a consistent pattern: \deepspeed exhibits latencies between \SI{177}{\milli\second} and \SI{180}{\milli\second}, whereas \sys processes the same layers in \SI{41.5}{\milli\second} to \SI{43.5}{\milli\second}, achieving up to $4.25 \times$ per-layer speedup.  

\subsection{TTFT of \sys and \deepspeed}
\label{sec:exp_speedup}

Next, we analyze how the \acf{TTFT} of \sys evolves with varying numbers of experts and batch sizes compared to \deepspeed. 
We define \ac{TTFT} as the time required for a full forward pass of the encoder. 
In these experiments, we fix the sequence length at 120, set the number of skewed experts ($k_r$) to 10\% of the total experts, and the skew degree ($\alpha_r$) to 0.6. 
We vary the number of experts from 8 to 256 with a fixed batch size of 250 and vary the batch size from 10 to 450 with 128 experts.

\begin{figure}[tpb]
    \centering
    \includegraphics{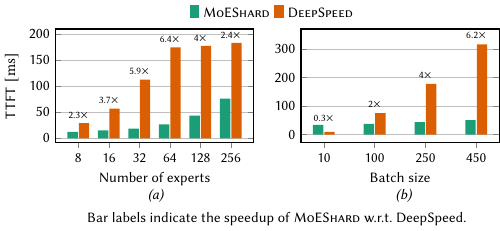}
    \caption{The average \ac{TTFT} of \sys with respect to \deepspeed for varying numbers of experts (left) and batch sizes (right). }
    \label{fig:speedup-exp}
\end{figure}

\Cref{fig:speedup-exp} compares \sys and \deepspeed under these settings. 
\Cref{fig:speedup-exp}a examines the impact of varying the number of experts, showing that \sys achieves increasing speedup, peaking at $6.45 \times$ before declining but remaining above $2.39 \times$. 
This decline results from \deepspeed's use of \ac{CF}, which drops tokens once the number of experts exceeds 50, reducing computation time.
In contrast, \sys remains consistently \emph{dropless}.
Overall, these results demonstrate that \sys outperforms \deepspeed in terms of \ac{TTFT} for different number of experts. %

\begin{figure}[tpb]
	\centering
	\includegraphics{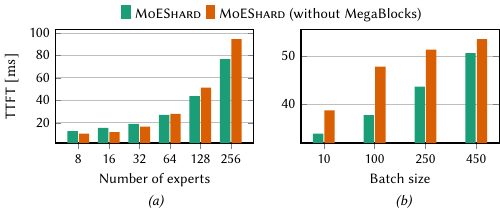}
	\caption{The average \ac{TTFT} of \sys with and without MegaBlocks enabled, for varying numbers of experts (left) and batch sizes (right).}
	\label{fig:ablation}
\end{figure}

\Cref{fig:speedup-exp}b presents results for varying batch sizes. Here, \sys is initially slower than \deepspeed at a batch size of 10, surpasses it at 100, and continues to show a near-linear increase in speedup, reaching approximately $6.24 \times$ at a batch size of 450. The lower initial speed of \sys is presumed to be due to fine-grained optimizations in \deepspeed that are absent in our implementation. However, the steady increase in speedup can be attributed to the fact that, even with fixed parameters of the custom router, as batch size grows, so does the absolute difference in token assignment across experts. This leads to greater total idle time for imbalanced solutions like \deepspeed.

\subsection{Ablation Study}
\label{sec:ablation_study}
Next, we break down the performance of \sys in its original formulation with the inclusion of Block Sparse Matrix Multiplication from \textsc{MegaBlocks}~\cite{megablocks}, as well as without it, referred to as \sys (without \textsc{MegaBlocks}).
We use the same experiment setup as in \Cref{sec:exp_speedup}.
The results, shown in~\Cref{fig:ablation}, reveal that for a fixed batch size and a varying number of experts, it is beneficial to exclude MegaBlocks until the number of experts reaches 64.
Beyond this point, the \textsc{MegaBlocks}-enhanced solution exhibits an increasingly higher speedup as the number of experts grows.
This can be explained by the overhead of kernel creation when using \textsc{MegaBlocks}; however, the benefits of \textsc{MegaBlocks} outweigh this overhead at higher expert counts, allowing it to fully exploit the advantages of \ac{TS}.
Furthermore, when varying the batch size, the solution with \textsc{MegaBlocks} consistently outperforms the one without it. This is attributed to the high number of experts (128) in this experiment, which allows \textsc{MegaBlocks} to achieve superior efficiency.

\section{Related Work}
\label{sec:related}

There is a vast body of work aimed at accelerating the inference of \ac{MoE} models across various areas, including improved load balancing, task scheduling, and communication optimization.
\textsc{DeepSpeed-MoE}~\cite{deepspeed-moe-paper} tackles multiple aspects of this challenge by providing a framework for serving \ac{MoE} models with highly optimized kernels, hierarchical all-to-all communication optimizations, and flexible parallelism strategies. 
However, its reliance on expert parallelism with static assignment makes it less effective for dynamic and unbalanced workloads. \textsc{Tutel}~\cite{hwang2023tuteladaptivemixtureofexpertsscale} improves on this by dynamically adjusting parallelism strategies at each iteration, yet it still struggles with load imbalance due to static expert assignment.
\textsc{Lazarus}~\cite{wu2024lazarus} addresses this issue using an optimal placement algorithm that replicates frequently selected experts across \acp{GPU} to better balance the workload at the cost of requiring additional \ac{GPU} memory. \textsc{Prophet}~\cite{wang2023prophet} instead builds a load-balancing placement model for experts and uses a greedy search to optimize their placement, while Lina~\cite{lina-paper} profiles experts and predicts their selection to enable dynamic resource scheduling.
Both systems, however, struggle when expert popularity shifts over time.
\textsc{ExFlow}~\cite{ex-flow-paper} takes a different approach by placing experts based on inter-layer affinity to reduce all-to-all communication and minimize token transfers between \acp{GPU}.
However, the system does not adapt well to affinity fluctuations caused by distribution change of inputs. %
In contrast to these approaches, \sys avoids complex scheduling mechanisms that may be sensitive to temporal shifts in expert popularity.
Instead, it places slices of experts across \acp{GPU}, significantly simplifying resource allocation.
Furthermore, \sys can take advantage of research on optimized CUDA kernels, enabling it to benefit from advancements in kernel optimization while ensuring efficient resource distribution.

\section{Conclusion}
\label{sec:conclusion}

In this paper, we presented \sys, a system that optimizes inference latency for \ac{MoE} models.
\sys ensures perfect load balancing across \acp{GPU} through tensor sharding of experts.
Our experiments demonstrate that \sys outperforms a state-of-the-art baseline across various settings for a high degree of routing function skewness.

\begin{acks}
This work has been funded by the Swiss National Science Foundation, under the project ``FRIDAY: Frugal, Privacy-Aware and Practical Decentralized Learning'', SNSF proposal No. 10.001.796.
\end{acks}

\bibliographystyle{unsrt}
\bibliography{main.bib}

\begin{thebibliography}{10}

\bibitem{fedus2022switchtransformers}
William Fedus, Barret Zoph, and Noam Shazeer.
\newblock Switch transformers: Scaling to trillion parameter models with simple
  and efficient sparsity.
\newblock {\em Journal of Machine Learning Research}, 23(120), 2022.

\bibitem{palm-model-paper}
Aakanksha Chowdhery, Sharan Narang, Jacob Devlin, Maarten Bosma, Gaurav Mishra,
  Adam Roberts, Paul Barham, Hyung~Won Chung, Charles Sutton, Sebastian
  Gehrmann, Parker Schuh, Kensen Shi, Sasha Tsvyashchenko, Joshua Maynez,
  Abhishek Rao, Parker Barnes, Yi~Tay, Noam Shazeer, Vinodkumar Prabhakaran,
  Emily Reif, Nan Du, Ben Hutchinson, Reiner Pope, James Bradbury, Jacob
  Austin, Michael Isard, Guy Gur-Ari, Pengcheng Yin, Toju Duke, Anselm
  Levskaya, Sanjay Ghemawat, Sunipa Dev, Henryk Michalewski, Xavier Garcia,
  Vedant Misra, Kevin Robinson, Liam Fedus, Denny Zhou, Daphne Ippolito, David
  Luan, Hyeontaek Lim, Barret Zoph, Alexander Spiridonov, Ryan Sepassi, David
  Dohan, Shivani Agrawal, Mark Omernick, Andrew~M. Dai,
  Thanumalayan~Sankaranarayana Pillai, Marie Pellat, Aitor Lewkowycz, Erica
  Moreira, Rewon Child, Oleksandr Polozov, Katherine Lee, Zongwei Zhou, Xuezhi
  Wang, Brennan Saeta, Mark Diaz, Orhan Firat, Michele Catasta, Jason Wei,
  Kathy Meier-Hellstern, Douglas Eck, Jeff Dean, Slav Petrov, and Noah Fiedel.
\newblock Palm: Scaling language modeling with pathways, 2022.

\bibitem{bommasani2021opportunities}
Rishi Bommasani, Drew~A Hudson, Ehsan Adeli, Russ Altman, Simran Arora, Sydney
  von Arx, Michael~S Bernstein, Jeannette Bohg, Antoine Bosselut, Emma
  Brunskill, et~al.
\newblock On the opportunities and risks of foundation models.
\newblock {\em arXiv preprint arXiv:2108.07258}, 2021.

\bibitem{environment-llms}
Matthias Rillig, Marlene Ågerstrand, Mohan Bi, Kenneth Gould, and Uli
  Sauerland.
\newblock Risks and benefits of large language models for the environment.
\newblock {\em Environmental science and technology}, 57, 02 2023.

\bibitem{moeoriginalpaper}
Noam Shazeer, Azalia Mirhoseini, Krzysztof Maziarz, Andy Davis, Quoc Le,
  Geoffrey Hinton, and Jeff Dean.
\newblock Outrageously large neural networks: The sparsely-gated
  mixture-of-experts layer, 2017.

\bibitem{singh2023hybrid}
Siddharth Singh, Olatunji Ruwase, Ammar~Ahmad Awan, Samyam Rajbhandari, Yuxiong
  He, and Abhinav Bhatele.
\newblock A hybrid tensor-expert-data parallelism approach to optimize
  mixture-of-experts training.
\newblock In {\em Proceedings of the 37th International Conference on
  Supercomputing}, pages 203--214, 2023.

\bibitem{fast-moe-paper}
Jiaao He, Jiezhong Qiu, Aohan Zeng, Zhilin Yang, Jidong Zhai, and Jie Tang.
\newblock Fastmoe: A fast mixture-of-expert training system, 2021.

\bibitem{faster-moe-paper}
Jiaao He, Jidong Zhai, Tiago Antunes, Haojie Wang, Fuwen Luo, Shangfeng Shi,
  and Qin Li.
\newblock Fastermoe: modeling and optimizing training of large-scale dynamic
  pre-trained models.
\newblock In {\em Proceedings of the 27th ACM SIGPLAN Symposium on Principles
  and Practice of Parallel Programming}, PPoPP '22, page 120–134, New York,
  NY, USA, 2022. Association for Computing Machinery.

\bibitem{smart-moe-paper}
Mingshu Zhai, Jiaao He, Zixuan Ma, Zan Zong, Runqing Zhang, and Jidong Zhai.
\newblock {SmartMoE}: Efficiently training {Sparsely-Activated} models through
  combining offline and online parallelization.
\newblock In {\em 2023 USENIX Annual Technical Conference (USENIX ATC 23)},
  pages 961--975, Boston, MA, July 2023. USENIX Association.

\bibitem{zhou2022expertchoicerouting}
Yanqi Zhou, Tao Lei, Hanxiao Liu, Nan Du, Yanping Huang, Vincent Zhao, Andrew~M
  Dai, zhifeng Chen, Quoc~V Le, and James Laudon.
\newblock Mixture-of-experts with expert choice routing.
\newblock In {\em Advances in Neural Information Processing Systems},
  volume~35. Curran Associates, Inc., 2022.

\bibitem{yang2021m6texploringsparseexpert}
An~Yang, Junyang Lin, Rui Men, Chang Zhou, Le~Jiang, Xianyan Jia, Ang Wang, Jie
  Zhang, Jiamang Wang, Yong Li, Di~Zhang, Wei Lin, Lin Qu, Jingren Zhou, and
  Hongxia Yang.
\newblock M6-t: Exploring sparse expert models and beyond, 2021.

\bibitem{kim2024scalingbeyondtheGPU}
Yechan Kim, Hwijoon Lim, and Dongsu Han.
\newblock Scaling beyond the {GPU} memory limit for large mixture-of-experts
  model training.
\newblock In {\em ICML}, 2024.

\bibitem{wang2023prophet}
Wei Wang, Zhiquan Lai, Shengwei Li, Weijie Liu, Keshi Ge, Yujie Liu, Ao~Shen,
  and Dongsheng Li.
\newblock Prophet: Fine-grained load balancing for parallel training of
  large-scale moe models.
\newblock In {\em 2023 IEEE International Conference on Cluster Computing
  (CLUSTER)}, pages 82--94. IEEE, 2023.

\bibitem{wu2024lazarus}
Yongji Wu, Wenjie Qu, Tianyang Tao, Zhuang Wang, Wei Bai, Zhuohao Li, Yuan
  Tian, Jiaheng Zhang, Matthew Lentz, and Danyang Zhuo.
\newblock Lazarus: Resilient and elastic training of mixture-of-experts models
  with adaptive expert placement.
\newblock {\em arXiv preprint arXiv:2407.04656}, 2024.

\bibitem{vaswani2017attention}
A~Vaswani et~al.
\newblock Attention is all you need.
\newblock {\em NeurIPS}, 2017.

\bibitem{shazeer2017outrageouslylargeneuralnetworks}
Noam Shazeer, Azalia Mirhoseini, Krzysztof Maziarz, Andy Davis, Quoc Le,
  Geoffrey Hinton, and Jeff Dean.
\newblock Outrageously large neural networks: The sparsely-gated
  mixture-of-experts layer.
\newblock In {\em ICLR}, 2017.

\bibitem{g-shard-paper}
Dmitry Lepikhin, HyoukJoong Lee, Yuanzhong Xu, Dehao Chen, Orhan Firat, Yanping
  Huang, Maxim Krikun, Noam Shazeer, and Zhifeng Chen.
\newblock Gshard: Scaling giant models with conditional computation and
  automatic sharding, 2020.

\bibitem{nvlink}
Ang Li, Shuaiwen~Leon Song, Jieyang Chen, Jiajia Li, Xu~Liu, Nathan~R. Tallent,
  and Kevin~J. Barker.
\newblock Evaluating modern gpu interconnect: Pcie, nvlink, nv-sli, nvswitch
  and gpudirect.
\newblock {\em IEEE Transactions on Parallel and Distributed Systems},
  31(1):94–110, January 2020.

\bibitem{megablocks}
Trevor Gale, Deepak Narayanan, Cliff Young, and Matei Zaharia.
\newblock Megablocks: Efficient sparse training with mixture-of-experts, 2022.

\bibitem{raffel2020exploring}
Colin Raffel, Noam Shazeer, Adam Roberts, Katherine Lee, Sharan Narang, Michael
  Matena, Yanqi Zhou, Wei Li, and Peter~J Liu.
\newblock Exploring the limits of transfer learning with a unified text-to-text
  transformer.
\newblock {\em Journal of machine learning research}, 21(140), 2020.

\bibitem{zhu2015moviebook}
Yukun Zhu, Ryan Kiros, Richard Zemel, Ruslan Salakhutdinov, Raquel Urtasun,
  Antonio Torralba, and Sanja Fidler.
\newblock Aligning books and movies: Towards story-like visual explanations by
  watching movies and reading books.
\newblock In {\em arXiv:1506.06724}, 2015.

\bibitem{deepspeed-moe-paper}
Samyam Rajbhandari, Conglong Li, Zhewei Yao, Minjia Zhang, Reza~Yazdani
  Aminabadi, Ammar~Ahmad Awan, Jeff Rasley, and Yuxiong He.
\newblock {D}eep{S}peed-{M}o{E}: Advancing mixture-of-experts inference and
  training to power next-generation {AI} scale.
\newblock In Kamalika Chaudhuri, Stefanie Jegelka, Le~Song, Csaba Szepesvari,
  Gang Niu, and Sivan Sabato, editors, {\em Proceedings of the 39th
  International Conference on Machine Learning}, volume 162 of {\em Proceedings
  of Machine Learning Research}, pages 18332--18346. PMLR, 17--23 Jul 2022.

\bibitem{hwang2023tuteladaptivemixtureofexpertsscale}
Changho Hwang, Wei Cui, Yifan Xiong, Ziyue Yang, Ze~Liu, Han Hu, Zilong Wang,
  Rafael Salas, Jithin Jose, Prabhat Ram, Joe Chau, Peng Cheng, Fan Yang, Mao
  Yang, and Yongqiang Xiong.
\newblock Tutel: Adaptive mixture-of-experts at scale, 2023.

\bibitem{lina-paper}
Jiamin Li, Yimin Jiang, Yibo Zhu, Cong Wang, and Hong Xu.
\newblock Accelerating distributed {MoE} training and inference with lina.
\newblock In {\em 2023 USENIX Annual Technical Conference (USENIX ATC 23)},
  pages 945--959, Boston, MA, July 2023. USENIX Association.

\bibitem{ex-flow-paper}
Jinghan Yao, Quentin Anthony, Aamir Shafi, Hari Subramoni, Dhabaleswar K., and
  Panda.
\newblock Exploiting inter-layer expert affinity for accelerating
  mixture-of-experts model inference, 2024.

\end{thebibliography}

\ifthenelse{\boolean{showcomments}}{

 }{}

\end{document}